\documentclass[conference]{IEEEtran}
\IEEEoverridecommandlockouts

\usepackage{cite}
\usepackage{amsmath,amssymb,amsfonts}
\usepackage{algorithmic}
\usepackage{graphicx}
\usepackage{textcomp}
\usepackage{xcolor}

\usepackage{cite}
\usepackage{amsmath,amssymb,amsfonts}
\usepackage{algorithmic}
\usepackage{graphicx}
\usepackage{textcomp}
\usepackage{xcolor}
\usepackage{multirow}
\usepackage{multirow}
\usepackage{makecell}
\usepackage{booktabs} 
\usepackage{multirow} 
\usepackage{amsmath}  

\usepackage{bbding}  
\usepackage{array}
\usepackage{hyperref}
\usepackage[normalem]{ulem} 

\def\BibTeX{{\rm B\kern-.05em{\sc i\kern-.025em b}\kern-.08em
    T\kern-.1667em\lower.7ex\hbox{E}\kern-.125emX}}
\begin{document}

\renewcommand{\thefootnote}{\arabic{footnote}}

\newcommand{\equalcontrib}{\textsuperscript{$^{*}$}}
\newcommand{\corrauthor}{\textsuperscript{\ddag}}

\title{Dental3R: Geometry-Aware Pairing for Intraoral 3D Reconstruction from Sparse-View Photographs%
\thanks{\noindent\parbox{\linewidth}{%
\equalcontrib~Co-first authors.\\
\corrauthor~Corresponding authors.}}%
}

\author{%
  \IEEEauthorblockN{%
    Yiyi Miao\textsuperscript{1,2}\equalcontrib,
    Taoyu Wu\textsuperscript{3,4,}\equalcontrib,
    Tong Chen\textsuperscript{1,2},
    Ji Jiang\textsuperscript{5},
    Zhe Tang\textsuperscript{6}, \\
    Zhengyong Jiang\textsuperscript{1},
    Angelos Stefanidis\textsuperscript{1},
    Limin Yu\textsuperscript{3}\corrauthor,
    Jionglong Su\textsuperscript{1}\corrauthor}
  \IEEEauthorblockA{%
    \textsuperscript{1}School of AI and Advanced Computing, Xi'an Jiaotong-Liverpool University, China\\
    \textsuperscript{2}School of Electrical Engineering, Electronics and Computer Science, University of Liverpool, United Kingdom\\
    \textsuperscript{3}School of Advanced Technology, Xi'an Jiaotong-Liverpool University, China\\
    \textsuperscript{4}School of Physical Sciences, University of Liverpool, Liverpool, United Kingdom\\
    \textsuperscript{5}School of Mathematics and Physics, Xi'an Jiaotong-Liverpool University, China\\
    \textsuperscript{6}Institute of Artificial Intelligence Innovation, Zhejiang University of Technology, China}
}

\maketitle
\begin{abstract}
Intraoral 3D reconstruction is fundamental to digital orthodontics, yet conventional methods like intraoral scanning are inaccessible for remote tele-orthodontics, which typically relies on sparse smartphone imagery. While 3D Gaussian Splatting (3DGS) shows promise for novel view synthesis, its application to the standard clinical triad of unposed anterior and bilateral buccal photographs is challenging. The large view baselines, inconsistent illumination, and specular surfaces common in intraoral settings can destabilize simultaneous pose and geometry estimation. Furthermore, sparse-view photometric supervision often induces a frequency bias, leading to over-smoothed reconstructions that lose critical diagnostic details. To address these limitations, we propose \textbf{Dental3R}, a pose-free, graph-guided pipeline for robust, high-fidelity reconstruction from sparse intraoral photographs. Our method first constructs a Geometry-Aware Pairing Strategy (GAPS) to intelligently select a compact subgraph of high-value image pairs. The GAPS focuses on correspondence matching, thereby improving the stability of the geometry initialization and reducing memory usage. Building on the recovered poses and point cloud, we train the 3DGS model with a wavelet-regularized objective. By enforcing band-limited fidelity using a discrete wavelet transform, our approach preserves fine enamel boundaries and interproximal edges while suppressing high-frequency artifacts. We validate our approach on a large-scale dataset of 950 clinical cases and an additional video-based test set of 195 cases. Experimental results demonstrate that Dental3R effectively handles sparse, unposed inputs and achieves superior novel view synthesis quality for dental occlusion visualization, outperforming state-of-the-art methods.
\end{abstract}


\section{Introduction}
Reliable intraoral 3D reconstruction underpins digital orthodontics: it informs diagnosis, treatment planning, appliance design, and longitudinal outcome assessment. While CBCT and intraoral scanning (IOS) deliver accurate 3D models and occlusal relationships, they require clinic-grade hardware, trained operators, and controlled environments, which limits accessibility for remote follow-ups and increases costs. In contrast, smartphone-assisted image capture has become common in tele-orthodontics; however, the resulting data are typically sparse, unposed, and of varying quality—making robust 3D reconstruction particularly challenging.

Recent progress in neural rendering and feed-forward geometry offers a promising alternative. 3D Gaussian Splatting (3DGS)~\cite{kerbl20233dgs} achieves photorealistic synthesis with efficient differentiable rasterization over explicit Gaussian primitives. To enhance robustness in challenging scenarios, a key trend is to solve for camera poses and scene structure simultaneously, thus bypassing the often brittle and slow SfM pre-processing step. CF-3DGS~\cite{cf3dgs} introduces an end-to-end pipeline for simultaneous camera tracking and novel view synthesis directly from video, eliminating any reliance on SfM pre-processing. DUSt3R~\cite{dust3r} regresses dense pointmaps from two unposed views, bypassing SfM initialization. However, directly coupling DUSt3R with 3DGS in intraoral settings exposes three practical failure modes: (i) dense or one-reference star pairing inflates memory/compute when propagating DUSt3R features across many pairs while still under-constraining large baselines; (ii) the clinical triad of frontal plus bilateral buccal images exhibits large view baselines, inconsistent illumination, and specular enamel, destabilizing camera recovery and radiance optimization and leading to drift, floating artifacts, and over-reconstruction under suboptimal pair selection; and (iii) with sparse viewpoints, purely photometric supervision induces frequency bias that smooths enamel edges and interproximal details, producing texture smearing even when geometry is roughly correct.

We present Dental3R, a pose-free, graph-guided pipeline tailored to sparse intraoral photographs. We construct a Hybrid Sequential Geometric View-Graph that arranges images on a cycle and proposes a bounded, multi-scale set of chords as pairing candidates. A range-aware importance model ranks candidate edges based on their expected geometric overlap, and a degree-bounded selection yields a compact subgraph that balances local reliability and global rigidity. Plugging this subgraph into DUSt3R focuses correspondence on high-value pairs, thereby improving pose stability and reducing memory usage. Building on the DUSt3R-initialized geometry, we train 3DGS with a wavelet-regularized objective. A two-level discrete wavelet transform enforces band-limited fidelity to preserve fine enamel boundaries and interproximal edges, while suppressing high-frequency artifacts under sparse views. Extensive experimental results demonstrate that Dental3R effectively handles sparse, unposed inputs and achieves superior novel view synthesis quality for dental occlusion visualization, outperforming state-of-the-art techniques.

\section{Related Work}

\subsection{Dense Input 3D Representations for Novel View Synthesis}
In past years, 3D reconstruction from images has been dominated by classical
pipelines combining SfM and Multi-View Stereo (MVS). SfM systems, such as the
widely-used COLMAP~\cite{colmap}, first recover a sparse 3D point cloud and
camera poses by matching local features across multiple views and performing
bundle adjustment. Subsequently, MVS algorithms densify this sparse
representation by leveraging photometric consistency across views. A paradigm
shift occurred with the introduction of Neural Radiance Fields
(NeRF)~\cite{mildenhall2021nerf}, which represents a scene as a continuous 5D
function learned by a Multi-Layer Perceptron (MLP). Given a dense set of input images with known camera poses, NeRF can produce high-fidelity new viewpoints, albeit with time-consuming per-scene training and rendering. However, the standard NeRF pipeline assumes rigid scenes and ample multi-view coverage, which are often violated in orthodontics, where anatomy is deformable and camera motion is limited. More recent approaches, such as 3D Gaussian Splatting(3DGS)~\cite{kerbl20233dgs,wu2025endoflow}, address NeRF’s efficiency limitations by representing scenes as explicit 3D Gaussian primitives. It initializes a set of 3D Gaussian splats from sparse SfM points and optimizes their color, opacity, and anisotropic covariance, thereby preserving the fidelity of volumetric fields without incurring heavy neural network inference in empty space. The result is dramatically faster training and rendering, making dense novel view synthesis more practical for interactive applications.

\subsection{Camera Pose-Free Reconstruction}
Another key trend has been lifting the requirement of known camera poses. Traditional pipelines run Structure-from-Motion (e.g. COLMAP) to estimate poses before reconstruction, but this can be slow or brittle for challenging scenes (e.g. texture-poor surgical images). Recent methods instead solve for camera poses simultaneously with scene reconstruction.
These methods aim to jointly optimize the scene representation and camera parameters. 
NoPe-NeRF~\cite{nopenerf} is a seminal method for optimizing Neural Radiance Fields without known camera poses. By leveraging monocular depth priors and introducing novel consistency losses, it jointly learns the scene representation and a consistent camera trajectory.
Compared to prior joint optimization methods (e.g. BARF~\cite{lin2021barf}, SC-NeRF~\cite{sc-nerf}) limited to forward-facing scenes, NoPe-NeRF can handle casual handheld videos, yielding both accurate novel view rendering and improved camera pose estimates. In parallel, researchers have integrated pose estimation into the 3DGS framework. 
CF-3DGS~\cite{cf3dgs} introduces an end-to-end pipeline for simultaneous camera tracking and novel view synthesis directly from video, eliminating any reliance on SfM pre-processing. The method progressively builds a global 3D Gaussian representation while estimating camera poses by registering this evolving model against each new incoming frame.
By eliminating reliance on COLMAP, CF-3DGS, and NoPe-NeRF, 3D reconstruction becomes more robust and accessible, a direction that is especially pertinent where calibrations or fiducial markers might otherwise be needed. Indeed, the push towards self-contained (pose-free) 3D vision is now informing surgical applications as well, as discussed next.

\subsection{Sparse Input Novel View Synthesis}
Capturing dozens of images per scene is impractical in many settings, motivating NVS approaches that work with sparse inputs~\cite{miao2025dentalsplat}. MVSNeRF~\cite{chen2021mvsnerf} pioneered fast generalizable radiance field reconstruction from as few as three views. By leveraging plane-sweep cost volumes (a multi-view stereo technique) to provide geometry cues, MVSNeRF’s network can infer a neural radiance field from just a handful of images. It achieves realistic novel views using minimal inputs and can be fine-tuned with additional images for higher quality, a significant speedup over the original NeRF, which required hours of per-scene optimization. More recently, MVSplat~\cite{mvsplat} introduces a feed-forward method that reconstructs a scene as a set of 3D Gaussians from a few wide-baseline images, using a plane-sweep cost volume to establish initial geometry in a single pass. Trained end-to-end with only photometric supervision, this approach achieves state-of-the-art novel view quality with significantly greater parameter efficiency, faster inference speed, and superior generalization compared to previous methods.

\subsection{End-to-End Reconstruction from Unposed Images}
DUSt3R~\cite{dust3r} introduced the idea of regressing dense 3D structure and camera poses directly from images in a feed-forward manner. DUSt3R predicts a pair of pointmaps given two input images, effectively solving pairwise Structure-from-Motion via neural regression. Building on this, MUSt3R~\cite{cabon2025must3r} extended the architecture to handle multi-view input by introducing a symmetric design and a latent memory that maintains a global frame of reference. In parallel, VGGT~\cite{wang2025vggt} introduces a versatile Transformer-based model that directly infers a comprehensive suite of 3D attributes, including camera parameters and scene geometry, from a widely variable number of input views. This approach achieves state-of-the-art results on multiple geometric tasks within a single forward pass, surpassing traditional pipelines by eliminating the need for iterative post-optimization, such as bundle adjustment.  Along the same lines, MapAnything~\cite{keetha2025mapanything} is a universal feed-forward 3D reconstruction model that unifies diverse tasks within one transformer-based architecture. MapAnything accepts flexible inputs – ranging from one or more images to additional cues such as intrinsics or partial reconstructions – and directly outputs the metric 3D scene structure and camera parameters.

\begin{figure*}[!t]
\centering
\includegraphics[width=1.0\textwidth]{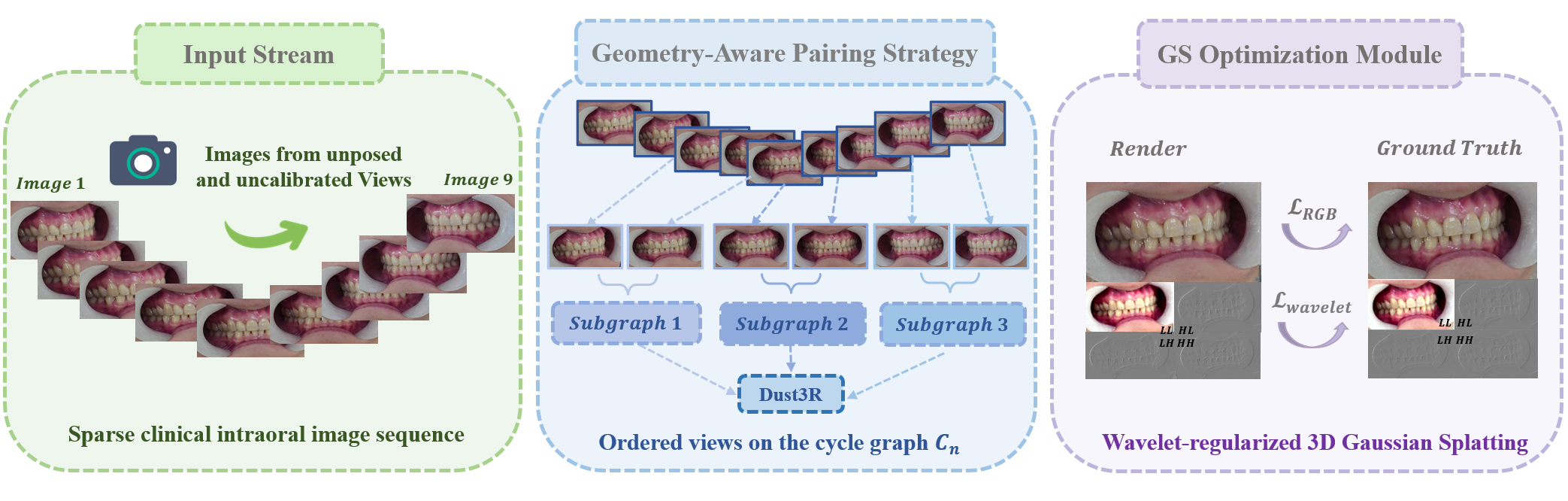}
\caption{{\textbf{Overview of Dental3R.} Given a set of sparse and unposed input images, we first employ our GAPS strategy to generate image pairs. Subsequently, we leverage a stereo-dense reconstruction model to regress a dense point cloud in a global coordinate system, while concurrently obtaining the corresponding relative camera poses. The resulting point cloud is then used to initialize the 3D Gaussians. During the optimization process, we incorporate wavelet constraints to ensure geometric consistency and frequency details.}}
\label{fig: 69viewNovel}
\end{figure*}

\section{Methodology}
\subsection{Preliminary}

\subsubsection{3D Gaussian Splatting.}
3DGS~\cite{kerbl20233dgs} is a high-fidelity radiance field method that represents a 3D scene as an explicit collection of anisotropic Gaussian primitives. Each Gaussian primitive $G_i$ is defined by a set of optimizable attributes: a center position $\boldsymbol{\mu}_i \in \mathbb{R}^3$, an opacity $o_i \in [0,1]$, a set of Spherical Harmonic (SH) coefficients for modeling view-dependent color, and a $3 \times 3$ covariance matrix $\boldsymbol{\Sigma}_i$. The spatial density of each Gaussian is given by:
\begin{equation}
  G_i(\mathbf{X}) = \exp\left\{-\frac{1}{2} (\mathbf{X} - \boldsymbol{\mu}_i)^\top \boldsymbol{\Sigma}_i^{-1} (\mathbf{X} - \boldsymbol{\mu}_i)\right\},
\end{equation}
where $\mathbf{X} \in \mathbb{R}^3$ denotes a point in 3D space. To ensure the covariance matrix remains positive semi-definite and to facilitate efficient optimization, $\boldsymbol{\Sigma}_i$ is parameterized by a 3D scaling vector $\mathbf{s}_i$ and a rotation quaternion $\mathbf{q}_i$.

The final color $\hat{C}(\mathbf{p})$ and depth $\hat{D}(\mathbf{p})$ for each pixel $\mathbf{p}$ are then synthesized by alpha-blending the contributions of all overlapping Gaussians, which are sorted from front to back along the camera ray. This volumetric rendering process is formulated as:
\begin{equation}
  \hat{C}(\mathbf{p}) = \sum_{i \in \mathcal{N}} c_i \alpha_i \prod_{j=1}^{i-1}(1-\alpha_j), \quad \hat{D}(\mathbf{p}) = \sum_{i \in \mathcal{N}} d_i \alpha_i \prod_{j=1}^{i-1}(1-\alpha_j),
\end{equation}
where $\mathcal{N}$ is the ordered set of Gaussians that along the pixel's viewing ray. For each Gaussian $i$, $c_i$ is the color derived from its SH coefficients based on the current viewing direction, and $d_i$ is its depth, corresponding to the z-coordinate of its center $\boldsymbol{\mu}_i$ in camera coordinates. 

\subsubsection{DUSt3R}
DUSt3R~\cite{dust3r} introduces a novel paradigm for dense 3D reconstruction from unconstrained image collections, operating without prior knowledge of camera intrinsic or extrinsic parameters. The framework circumvents traditional Structure-from-Motion (SfM) pipelines by directly regressing a dense scene representation called a pointmap. A pointmap, denoted as $\mathbf{X} \in \mathbb{R}^{W \times H \times 3}$, establishes a direct mapping from the pixel coordinates $(u,v)$ of an image $\mathbf{I}$ to 3D points in a specific coordinate system. Given a depth map $\mathbf{D}$ and camera intrinsics $\mathbf{K}$, the pointmap is defined as:
\begin{equation}
  \mathbf{X}_{u,v} = D_{u,v} \cdot \mathbf{K}^{-1} \begin{bmatrix} u, v, 1 \end{bmatrix}^T,
\end{equation}
where $\mathbf{X}$ is expressed in the camera's local coordinate frame. The core network architecture takes a pair of images, $\mathbf{I}_1$ and $\mathbf{I}_2$, and outputs two corresponding pointmaps, $\mathbf{X}^{1,1}$ and $\mathbf{X}^{2,1}$, that are implicitly aligned to a common reference frame.

The model is trained using a 3D regression loss designed to minimize the Euclidean distance between predicted and ground-truth pointmaps. To address the inherent scale ambiguity in uncalibrated stereo reconstruction, both the predictions and the ground truth are normalized. The normalized regression loss for a pixel $i$ in view $v$ is formulated as:
\begin{equation}
  \mathcal{L}_{\text{reg}}(v,i) = \left\| \frac{1}{s} \mathbf{X}_{i}^{v,1} - \frac{1}{\bar{s}} \bar{\mathbf{X}}_{i}^{v,1} \right\|,
\end{equation}
where $\mathbf{X}$ and $\bar{\mathbf{X}}$ are the predicted and ground-truth pointmaps, respectively. The scale factors, $s$ and $\bar{s}$, are computed as the mean distance of all valid 3D points from the origin, ensuring scale-invariant comparison.

Furthermore, to handle regions that are inherently difficult to reconstruct, such as textureless surfaces, sky, or transparent objects, DUSt3R employs a confidence-aware training objective. The network jointly predicts a per-pixel confidence map $\mathbf{C}$, which modulates the regression loss. The final confidence-aware loss function is:
\begin{equation}
  \mathcal{L}_{\text{conf}} = \sum_{v \in \{1,2\}} \sum_{i \in \mathcal{D}^v} \left( C_{i}^{v,1} \mathcal{L}_{\text{reg}}(v,i) - \alpha \log C_{i}^{v,1} \right),
  \label{eq:conf_loss}
\end{equation}
where $\mathcal{D}^v$ is the set of valid pixels in view $v$, and the logarithmic term regularizes the confidence prediction~. This objective enhances robustness against geometric ambiguities and yields a per-pixel confidence that is valuable for downstream tasks.

\begin{figure*}[!t]
\centering
\includegraphics[width=0.9\textwidth]{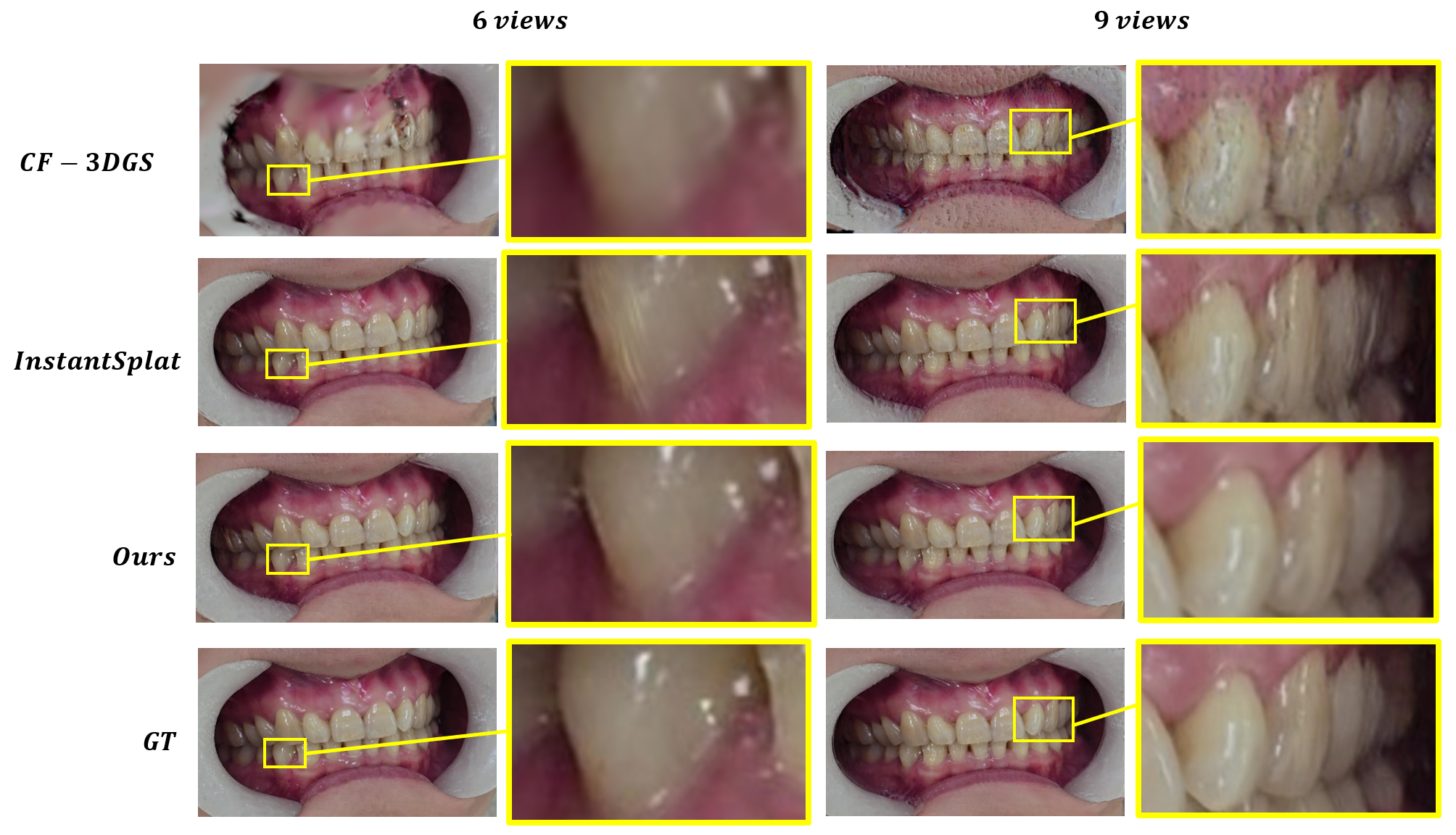}
\caption{{\textbf{Novel View Synthesis Comparisons with 6 views and 9 views input.}
We qualitatively compare the quality of novel view synthesis with 3DGS~\cite{kerbl20233dgs}, CF-3DGS~\cite{cf3dgs}, and InstantSplat~\cite{instantsplat}, and show that our method achieves better quality and more accurate texture details.}}
\label{fig: 69viewNovel}
\end{figure*}

\begin{table*}[htb]
  \centering
  \caption{Quantitative evaluation on the video test set, using different input viewpoints. The best results are bolded.}
  \label{tab:6_9view}
  \setlength{\tabcolsep}{6pt} %
  \begin{tabular}{l ccc ccc ccc ccc}
    \toprule
    \multirow{2}{*}{Algorithm} &
    \multicolumn{3}{c}{3 Training views} & 
    \multicolumn{3}{c}{6 Training views} &
    \multicolumn{3}{c}{9 Training views} &
    \multicolumn{3}{c}{12 Training views} \\
    \cmidrule(lr){2-4}\cmidrule(lr){5-7}\cmidrule(lr){8-10}\cmidrule(lr){11-13}
    & PSNR$\uparrow$ & SSIM$\uparrow$ & LPIPS$\downarrow$
    & PSNR$\uparrow$ & SSIM$\uparrow$ & LPIPS$\downarrow$
    & PSNR$\uparrow$ & SSIM$\uparrow$ & LPIPS$\downarrow$
    & PSNR$\uparrow$ & SSIM$\uparrow$ & LPIPS$\downarrow$ \\
    \midrule
    3DGS~\cite{kerbl20233dgs}        & --    & --     & --     & --     & --     & --     & --     & --     & --     & 11.51  & 0.53  & 0.574 \\
    InstantSplat~\cite{instantsplat} & 23.81 & \textbf{0.826} & 0.304  & 27.01  & 0.863  & 0.268  & 28.66 & 0.890  & 0.241  & 29.32 & 0.898 & 0.235 \\
    CF-3DGS~\cite{cf3dgs}            & 15.32 & 0.748  & 0.443  & 18.01  & 0.795  & 0.277  & 21.29  & 0.812  & 0.374  & 23.02  & 0.853 & 0.337 \\
    \textbf{Ours}                    & \textbf{23.87} & 0.824  & \textbf{0.300} & \textbf{27.97} & \textbf{0.869} & \textbf{0.251} & \textbf{29.161} & \textbf{0.892} & \textbf{0.237} & \textbf{29.947} & 0.884 & \textbf{0.221} \\
    \bottomrule
  \end{tabular}
  \vspace{-0.5em}
\end{table*}

\subsection{Hybrid Sequential Geometric Scene Graph}

\paragraph{Motivation}
In multi-view 3D reconstruction, the selection of image pairs is a critical determinant of both computational efficiency and model accuracy. Dense or complete graphs (pairing every view) are computationally prohibitive for large datasets and, when used with modern Transformer-based priors such as DUSt3R, consume large amounts of memory and require high-performance GPUs, making them unsuitable for practical medical scenarios. Conversely, overly sparse graphs or suboptimal strategies like the oneref method from DUST3R may lack sufficient geometric constraints, leading to inaccurate camera poses and a degradation in the final reconstruction quality.

Therefore, to address these limitations, we propose the Geometry-Aware Pairing Strategy (GAPS) to construct a sparse measurement graph that balances local reliability and global rigidity. GAPS operates on a cycle of views, defining multi-scale neighborhoods to efficiently enforce both robust local and global constraints, thereby ensuring accurate modeling. We formalize this by letting the view indices form the vertex set $V=\{0,1,\dots,n-1\}$ with $n\in\mathbb{N}$, arranged on the cycle graph $C_n$. A small, fixed set of positive offsets $\mathcal{K}$ encodes a multi-scale neighborhood on $C_n$. Throughout, $\{i,j\}$ denotes an undirected edge between vertices $i,j\in V$, and all index arithmetic is taken modulo $n$.

\paragraph{Sequential Prior and Candidate Chords}
Multi-scale sequential structure proposes a bounded set of chords as candidates:
\begin{equation}
E_{\mathrm{cand}}
\,=\,
\bigcup_{k\in\mathcal{K}}
\bigl\{\{i,\,(i+k)\bmod n\}\;:\; i\in V\bigr\},
\label{eq:candidates}
\end{equation}
where \(E_{\mathrm{cand}}\) is the candidate edge set obtained by uniting the \(k\)-ring chords on \(C_n\) across \(k\in\mathcal{K}\).

\paragraph{Geometric Overlap and Edge Importance}
Expected image overlap is modeled by a wrap-around index distance on the cycle:
\begin{equation}
d(i,j) \,=\, \min\!\bigl(|i-j|,\; n-|i-j|\bigr),
\label{eq:distance}
\end{equation}
where \(d:V\times V\to \{0,1,\dots,\lfloor n/2\rfloor\}\) is the circular distance on \(C_n\) and \(\lfloor\cdot\rfloor\) denotes the floor operator. 

A monotone decay \(\phi\) maps this distance to a surrogate overlap, which is converted to a range-adaptive importance:
\begin{equation}
w(i,j) \,=\, \alpha_{r(i,j)}\,\phi\!\bigl(d(i,j)\bigr)\;+\;\beta_{r(i,j)} ,
\label{eq:weight}
\end{equation}
where \(w:E_{\mathrm{cand}}\to\mathbb{R}_{\ge 0}\) is the edge-importance function, \(\phi:[0,\lfloor n/2\rfloor]\to[0,1]\) is a fixed monotone decay, \(r(i,j)\in\{\text{local},\text{medium},\text{long}\}\) is a coarse range class derived from \(d(i,j)\), and \(\alpha_{\cdot},\beta_{\cdot}\in\mathbb{R}\) are range-dependent constants chosen once. Edges with low importance are discarded, yielding a weighted candidate graph.

\paragraph{Degree-Bounded High-Weight Subgraph}
To control complexity and prevent hubs while favoring informative connections, we seek a degree-constrained, high-weight subgraph:
\begin{equation}
\max_{E'\subseteq E_{\mathrm{cand}}}\;\sum_{\{i,j\}\in E'} w(i,j)
\quad\text{s.t.}\quad
\deg_{E'}(v)\le b\;\;\forall v\in V,
\label{eq:bmatching}
\end{equation}
where \(E'\) is the selected edge set, \(\deg_{E'}(v)\) is the degree of vertex \(v\) in \((V,E')\), and \(b\in\mathbb{N}\) is a uniform degree cap. 

By adopting the proposed pairing strategy in the initial stage of Dust3R~\cite{dust3r}, we effectively reduced memory usage while maintaining training results.

\subsection{Wavelet Decomposition}

Wavelet analysis provides a mathematical framework for the multi-resolution representation of signals. Its principal advantage over Fourier analysis is time-frequency localization~\cite{wu2025endowave}. Unlike the Fourier transform, which utilizes globally supported sinusoidal basis functions, the wavelet transform employs basis functions that are localized in both the time and frequency domains. For two-dimensional signals, such as images, this property enables a decomposition into sub-bands of varying spatial frequency and orientation, while preserving crucial spatial information.

The two-dimensional Discrete Wavelet Transform (DWT) is applied to a discrete image, represented as a matrix $I \in \mathbb{R}^{H \times W}$. The transform utilizes a pair of Quadrature Mirror Filters (QMFs), a low-pass filter $h$ and a high-pass filter $g$, which are applied separably to the image's rows and columns. Each filtering stage is followed by dyadic downsampling. This process yields four coefficient sub-bands:
\begin{equation}
\begin{aligned}
LL &= (I *_{\text{row}} g) *_{\text{col}} g \\
LH &= (I *_{\text{row}} g) *_{\text{col}} h \\
HL &= (I *_{\text{row}} h) *_{\text{col}} g \\
HH &= (I *_{\text{row}} h) *_{\text{col}} h
\end{aligned}
\label{eq:dwt_final}
\end{equation}
Here, $*_{\text{row}}$ and $*_{\text{col}}$ denote one-dimensional convolution along the respective image dimensions. The resulting coefficient maps represent the approximation ($LL$), horizontal details ($LH$), vertical details ($HL$), and diagonal details ($HH$).

Our framework incorporates a wavelet loss term to regularize the optimization process in the frequency domain, based on a two-level DWT. First, we define the residual map $\Delta_x$ for each frequency component $x$ as the difference between the ground truth image $I_\text{gt}$ and the rendered output $\hat{I}$:
\begin{equation}
    \Delta_x = W_x(I_\text{gt}) - W_x(\hat{I}),
\end{equation}
where $W_x$ denotes the operation that extracts the sub-band $x \in \{LL, LH, HL, HH\}$. The total wavelet loss is then computed as the weighted sum of the squared $L_2$ norms of these residual maps:
\begin{equation}
    \mathcal{L}_\text{wavelet} = \sum_{x \in \{LL,LH,HL,HH\}} \lambda_x \left\| \Delta_x \right\|_2^2.
\end{equation}
The weights $\lambda_x$ control the influence of each frequency component on the wavelet loss.

\section{Experiment}

\begin{table*}[!t]
  \centering
  \caption{Ablation study and pairing-strategy comparison on the video test set. Best is in \textbf{bold}, the second best is \uline{underlined}.}
  \label{tab:views_split}
  \setlength{\tabcolsep}{6pt}
  \renewcommand{\arraystretch}{1.1}

  \resizebox{\textwidth}{!}{%
  \begin{tabular}{l ccccc ccccc}
    \toprule
    \multirow{2}{*}{Method} &
    \multicolumn{5}{c}{3 Training views} &
    \multicolumn{5}{c}{6 Training views} \\
    \cmidrule(lr){2-6}\cmidrule(lr){7-11}
    & Pairs & SSIM$\uparrow$ & PSNR$\uparrow$ & LPIPS$\downarrow$ & GPU Memory (MB)$\downarrow$
    & Pairs & SSIM$\uparrow$ & PSNR$\uparrow$ & LPIPS$\downarrow$ & GPU Memory (MB)$\downarrow$ \\
    \midrule
    Cosine~\cite{gao2025easysplat}  &  2 &  0.762 & 20.20 & 0.401 & 3746
                                    & 10 & 0.809  & 22.817 & 0.360 & \textbf{2984} \\
    Complete~\cite{instantsplat}    &  6 & \textbf{0.834} & \textbf{25.65} & \textbf{0.286} & 7848
                                    & 30 & \textbf{0.873} & \textbf{28.77} & \textbf{0.247} & 10158 \\
    Oneref~\cite{instantsplat}            &  4 & 0.790  & 24.01 & 0.340 & 3542
                                    & 10 & 0.839  & 26.72 & 0.295 & \uline{4710} \\
    {Ours w.o.\ Wavelet}           &  6 & 0.819  & 24.41 & 0.324 & \textbf{2414}
                                    &  9 & 0.857  & 28.27 & \uline{0.264} & 6479 \\
    \textbf{Ours}                   &  6 & \uline{0.820} & \uline{24.55} & \uline{0.323} & \uline{2744}
                                    &  9 & \uline{0.860} & \uline{28.35} & \uline{0.264} & 6420 \\
  \end{tabular}%
  }

  \vspace{0.8em}

  \resizebox{\textwidth}{!}{%
  \begin{tabular}{l ccccc ccccc}
    \toprule
    \multirow{2}{*}{Method} &
    \multicolumn{5}{c}{9 Training views} &
    \multicolumn{5}{c}{12 Training views} \\
    \cmidrule(lr){2-6}\cmidrule(lr){7-11}
    & Pairs & SSIM$\uparrow$ & PSNR$\uparrow$ & LPIPS$\downarrow$ & GPU Memory (MB)$\downarrow$
    & Pairs & SSIM$\uparrow$ & PSNR$\uparrow$ & LPIPS$\downarrow$ & GPU Memory (MB)$\downarrow$ \\
    \midrule
    Cosine~\cite{gao2025easysplat}  & 16 & 0.819 & 23.17 & 0.324 & 4144
                                    & 22 & 0.871 & 25.46 & 0.275 & 4320 \\
    Complete~\cite{instantsplat}    & 72 & \uline{0.893} & 29.26 & \textbf{0.229} & 12888
                                    & 132 & \textbf{0.904} & 29.76 & \uline{0.224} & 16038 \\
    Oneref~\cite{instantsplat}            & 16 & 0.871 & 28.27 & 0.258 & \uline{4880}
                                    &  22 & 0.886 & 29.27 & 0.241 & \uline{5068} \\
    {Ours w.o.\ Wavelet}           & 27 & 0.892 & \uline{29.61} & 0.232 & 7427
                                    &  38 & 0.899 & \uline{29.81} & 0.227 & 7664 \\
    \textbf{Ours}                   & 27 & \textbf{0.894} & \textbf{29.62} & \uline{0.231} & 7636
                                    &  38 & \uline{0.900} & \textbf{29.84} & \textbf{0.221} & 7623 \\
    \bottomrule
  \end{tabular}%
  }
    \label{tab:ablation_results}
  \vspace{-0.25em}
\end{table*}

\subsection{Implementation Details}
\subsubsection{Dataset Description}

To rigorously assess accuracy and robustness, we assembled a clinical intraoral dataset in partnership with specialist dental hospitals. All imagery was acquired by certified orthodontists using a Canon EOS~700D with a 100\,mm macro lens under forced-flash settings, ensuring consistent illumination and minimizing lighting variability. The dataset comprises two subsets. \textbf{(i) Video subset:} 195 clinical cases, each a short intraoral sequence spanning a continuous sweep from right buccal to frontal occlusion to left buccal. From every sequence we uniformly sampled 24 frames, then split them evenly into 12 training and 12 test images. Training was performed using only the training images and their corresponding camera poses. We evaluated four sparse-view regimes with 3, 6, 9, and 12 input views. After optimization, novel views were rendered at the test poses to assess the quality of synthesis. \textbf{(ii) Three-image subset:} 950 clinical cases, each containing exactly three photographs (anterior occlusal, left buccal, right buccal), used to probe reconstruction and novel-view synthesis under extremely sparse observations.

\subsubsection{Experimental Setup}
All experiments were conducted on a desktop workstation equipped with an Intel Core i9-13900KF processor and an NVIDIA GeForce RTX 4090 GPU. To ensure a fair and consistent comparison across all cases, a uniform set of hyperparameters was applied throughout the experiments. For the optimization of the 3D Gaussian attributes, we adhered to the default training parameters established in the original Gaussian Splatting implementation~\cite{kerbl20233dgs}. The parameters were updated using the Adam optimizer~\cite{adam}. To balance rendering efficiency and quality, we set the number of training iterations to 3000.

\subsection{Evaluation results}

\begin{figure*}[!t]
\centering
\includegraphics[width=\textwidth]{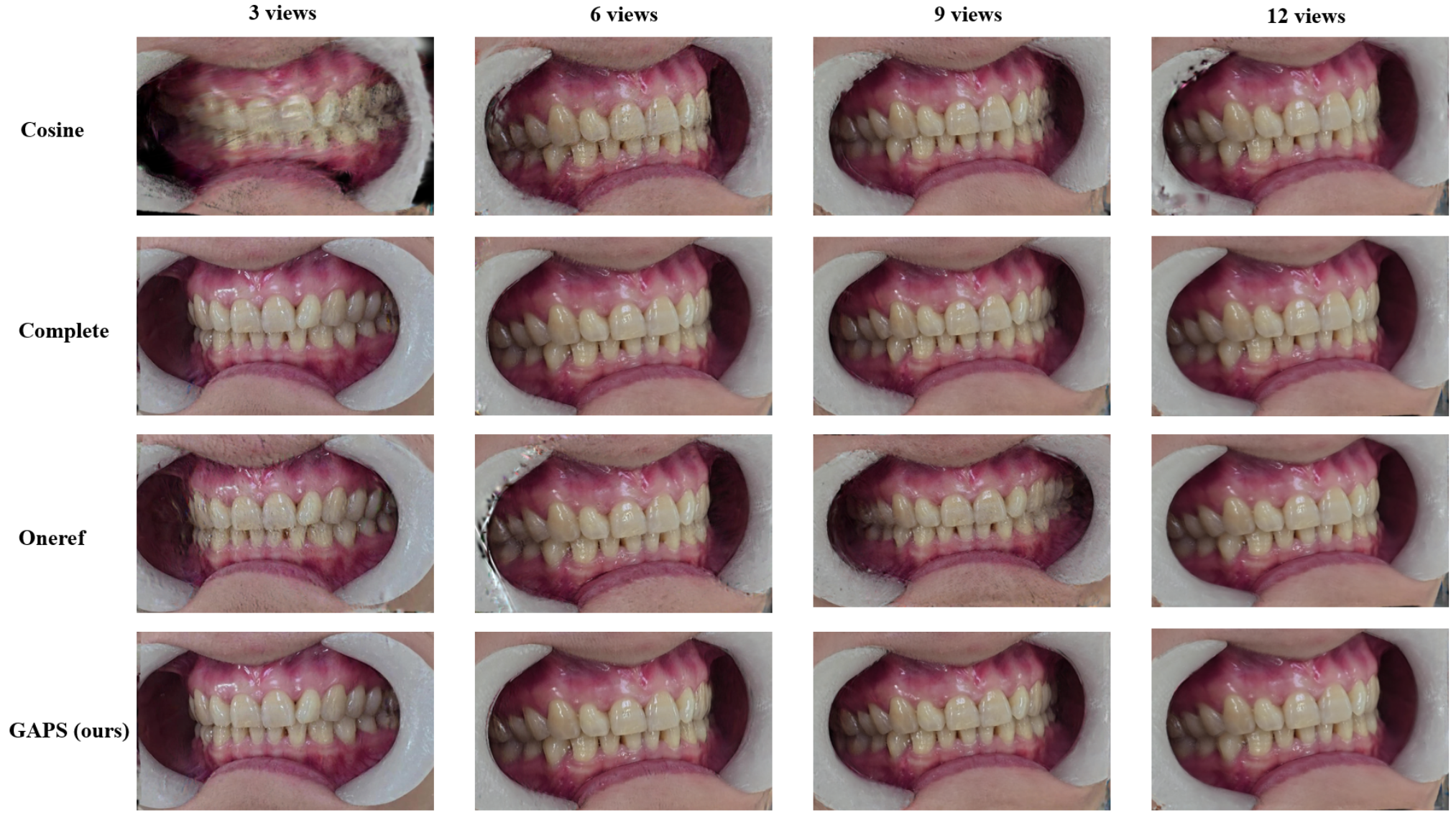}
\caption{{\textbf{Novel View Synthesis Results with Different Pair Strategy.}
We perform a qualitative comparison against the complete and oneref graph strategies from InstantSplat~\cite{instantsplat}, as well as the cosine graph strategy from EasySplat~\cite{gao2025easysplat}. The results demonstrate that our proposed GAPS strategy achieves novel view synthesis performance competitive with the exhaustive complete strategy. Furthermore, GAPS surpasses the other two sparse methods (oneref and cosine), yielding superior rendering quality and more accurate textural details across various input views.}}
\label{fig: paircompare}
\end{figure*}

\subsubsection{Comparative Experiments on Video Test Dataset}

To assess our framework's novel view synthesis capabilities, we benchmarked it against the original 3DGS and other state-of-the-art methods, including CF-3DGS and InstantSplat, using our comprehensive video dataset. The quantitative results, summarized in Table~\ref{tab:6_9view}. Averaged across 195 distinct cases, our method consistently outperforms all baselines across the three standard metrics: Peak Signal-to-Noise Ratio (PSNR)~\cite{psnr}, Structural Similarity Index Measure (SSIM)~\cite{SSIM}, and Learned Perceptual Image Patch Similarity (LPIPS)~\cite{Lpips}.

Notably, the performance of standard 3DGS is of particular interest. The model fails to converge entirely when initialized with sparse multi-view inputs, as indicated by the "-" entries in the table. Even in the more data-rich 12-view setting, where it is trainable, its rendering quality remains substantially inferior. This behavior highlights a fundamental limitation of conventional approaches in sparse-observation scenarios, which are typical of dental imaging.

Figure~\ref{fig: 69viewNovel} illustrates the qualitative evaluations of novel view synthesis. Under the challenging 6-view condition, outputs from CF-3DGS are plagued by noticeable blurring and floating artifacts, while InstantSplat exhibits pronounced geometric distortions in the lower teeth. Although increasing the inputs to 9 views allows CF-3DGS to mitigate major artifacts, the resulting images are still of low resolution and suffer from clear overfitting, such as hallucinated geometry in the right buccal area. InstantSplat also struggles with over-reconstruction in the right molars, leading to texture and shape degradation. Such artifacts could critically undermine the reliability of clinical assessments, particularly in applications such as remote orthodontic monitoring. In stark contrast, our reconstructions maintain exceptional geometric fidelity and remain artifact-free across both input conditions, demonstrating robust generalization for high-quality novel view synthesis.

\subsubsection{Comparative Experiments on 3 Views images Dataset}
\begin{table}[h]
  \centering
  \caption{\textbf{Quantitative results} with the other methods using 3 views.}
  \begin{tabular}{c c c c c}
    \hline
    Methods                          & PSNR$\uparrow$ & SSIM$\uparrow$ & LPIPS$\downarrow$ & Times (Seconds)$\downarrow$ \\
    \hline
    3DGS~\cite{kerbl20233dgs}        & -              & -              & -                 & -                           \\
    InstantSplat~\cite{instantsplat} & 32.78          & 0.945          & 0.160             & \textbf{57}                 \\
    CF-3DGS~\cite{cf3dgs}            & 18.37          & 0.803          & 0.32              & 372                         \\
    \textbf{Ours}                    & \textbf{33.89} & \textbf{0.949} & \textbf{0.137}    & 64                          \\
    \hline
  \end{tabular}
  \label{tab:3view}
\end{table}

The framework's robustness was further challenged under extremely sparse input conditions using a large-scale image dataset of 950 clinical cases. For this experiment, occlusal reconstruction was performed from a minimal set of three intra-oral images: the anterior, left buccal, and right buccal views. Given the absence of novel test views in this dataset, our quantitative evaluation in Table~\ref{tab:3view} assesses reconstruction fidelity on the training views themselves, averaged over 956 cases.

\subsection{Ablation Study}

To validate the contribution of each key component in our framework, we conduct an ablation study on the proposed graph strategy GAPS and the wavelet constraint, with results summarized in Table~\ref{tab:ablation_results}. We compare the Complete and oneref graph strategy from InstantSplat~\cite{instantsplat} and the cosine graph strategy from EasySplat~\cite{gao2025easysplat}, as shown in the Figure~\ref{fig: paircompare}. The Ours w.o.\ Wavelet shows the results of our graph strategy with the wavelet constraint removed. The analysis first highlights the efficacy of our graph strategy. We observe that it produces image pairs that are more amenable to optimization, which is crucial for achieving high-quality novel view synthesis. 
Moreover, this strategy achieves significantly lower GPU memory usage than the Complete graph strategy baseline in InstantSplat~\cite{instantsplat}, while maintaining comparable performance. The quantitative results in the table, averaged over 20 randomly sampled cases from the video test dataset. The best result is bolded, and the second result is underlined, showing the effectiveness and efficiency of our work.

\section{Conclusion}
In this paper, we introduced Dental3R, a novel, graph-guided pipeline designed for high-fidelity 3D reconstruction of dental occlusion from sparse and unposed intraoral photographs. We addressed the challenges in the tele-orthodontic scene: unstable camera pose estimation arising from large view baselines and the loss of fine diagnostic details due to photometric supervision under sparse views. Our proposed GAPS strategy stabilizes the reconstruction process by intelligently selecting optimal image pairs for a robust and efficient geometry initialization. Furthermore, our wavelet-regularized objective for 3DGS training effectively counteracts frequency bias, preserving the sharp enamel edges and interproximal details that are essential for clinical assessment. Extensive experiments on our large-scale clinical dataset demonstrate that Dental3R significantly outperforms state-of-the-art models in terms of novel view synthesis quality.


\bibliographystyle{IEEEtran}
\bibliography{dentalBIBM}

\end{document}